\documentclass{article}

\usepackage{microtype}
\usepackage{graphicx}
\usepackage{subcaption}
\usepackage{booktabs} 
\usepackage{breakurl}
\usepackage{enumitem}
\usepackage{scalerel}
\usepackage{pifont}
\usepackage{tikz}
\usepackage{tabularx}
\usepackage{multirow}

\newcolumntype{Y}{>{\raggedright\arraybackslash}X}

\usepackage[hyphens]{url}
\usepackage{hyperref}

\usepackage[accepted]{icml2026}

\usepackage{amsmath}
\usepackage{amssymb}
\usepackage{mathtools}
\usepackage{amsthm}
\usepackage{tcolorbox}

\usepackage[capitalize,noabbrev]{cleveref}

\theoremstyle{plain}

\theoremstyle{definition}

\theoremstyle{remark}

\usepackage{diagbox}
\usepackage{makecell}
\usepackage{float}
\usepackage{threeparttable}
\usepackage{placeins}
\usepackage{listings}
\usepackage{ebproof}

\usepackage[disable,textsize=tiny]{todonotes}

\definecolor{red}{RGB}{220,40,30}
\definecolor{BrightOrange}{RGB}{255,140,0}
\definecolor{Gold}{RGB}{255,200,0}
\definecolor{Emerald}{RGB}{0,160,80}
\definecolor{Cobalt}{RGB}{0,120,215}
\definecolor{Sapphire}{RGB}{90,60,170}
\definecolor{Amethyst}{RGB}{160,80,200}
\definecolor{framegray}{RGB}{170,170,170}
\definecolor{mygray}{RGB}{242,242,242}
\icmltitlerunning{Position: Multi-Agent Systems Should Prioritize Concurrency Control}

\begin{document}

\pagecolor{mygray} 

\twocolumn[
  \icmltitle{Position: Multi-Agent Systems Should Prioritize Concurrency Control}



  \icmlsetsymbol{equal}{*}
  \icmlsetsymbol{cores}{$\dagger$}

  \begin{icmlauthorlist}
    \icmlauthor{Xin Yang}{equal,zju}
    \icmlauthor{Letian Li}{equal,tbsi}
    \icmlauthor{Zimo Ji}{hkust}
    \icmlauthor{Terry Jingchen Zhang}{cores,eth}
    \icmlauthor{Wenyuan Jiang}{cores,eth}
  \end{icmlauthorlist}

  \icmlaffiliation{tbsi}{Shenzhen International Graduate School, Tsinghua University, Shenzhen, China}
  \icmlaffiliation{zju}{School of Mathematical Sciences, Zhejiang University, Hangzhou, China}
  \icmlaffiliation{eth}{ETH Zurich, Zurich, Switzerland}
  \icmlaffiliation{hkust}{Hong Kong University of Science and Technology, Clear Water Bay, Hong Kong}

  \icmlcorrespondingauthor{Wenyuan Jiang}{wenyjiang@ethz.ch}

  \icmlkeywords{Machine Learning, ICML}

  \vskip 0.3in
]



\printAffiliationsAndNotice{\icmlEqualContribution}

\begin{abstract}
LLM-based multi-agent systems (MAS) promise scalable collaboration, yet adding agents often \emph{reduces} reliability. 
This position paper argues that many MAS failures are fundamentally \textbf{concurrency control problems}: agents concurrently read and write shared state, and long LLM inference windows amplify the risk of stale reads, lost updates, and inconsistent outcomes. 
Failure modes commonly attributed to ``coordination'' or ``communication'' breakdowns can be mapped directly onto classical concurrency anomalies. 
We contend that MAS frameworks should address these failures through explicit concurrency control mechanisms: conflict detection, isolation guarantees, and structured access to shared resources.
Concurrency control should be a first-class design concern, not an afterthought.
\end{abstract}

\section{Introduction}
\label{sec:intro}

Large language models (LLMs) have demonstrated increasingly powerful capabilities in language understanding, reasoning, and generation \citep{brown2020language,openai2024gpt4technicalreport,touvron2023llama,yang2026towards}, expanding their application domains from text generation to complex decision-making. 
The integration of tool calling and environmental interaction has further transformed LLMs into autonomous agents capable of executing real-world tasks through iterative reasoning and action \citep{yao2023react,schick2023toolformer,mialon2023augmented,cheng2025higher}. 
This paradigm shift, exemplified by the ReAct framework \citep{yao2023react}, enables models to move beyond pure dialogue toward practical task completion through observation, reasoning, and action loops. 
More recently, the pursuit of solving increasingly complex tasks has driven the development of multi-agent systems (MAS), where multiple agents operate \emph{concurrently}, communicating and collaborating to achieve goals at greater scale and with higher success rates than single-agent approaches \citep{park2023generative,li2023camel,hong2024metagpt,qian2024chatdev,wu2024autogen,guo2024large,icml2026sub}. 
However, empirical evidence reveals a sobering reality: scaling the number of agents does not reliably improve performance \citep{cemri2025multi,kim2025towards}. 
Studies show that multi-agent systems exhibit failure rates between 41\% and 86.7\% across popular benchmarks, with coordination failures and inter-agent misalignment accounting for a substantial fraction of these breakdowns \citep{cemri2025multi,li2024more}. 
Recent systems and benchmarks provide concrete evidence that concurrency mechanisms affect MAS outcomes \citep{geng2026effective,chen2024coder,zhang2026silo,wang2024megaagent,chang2025sagallm}.

\begin{figure}[t]
\centering
\includegraphics[width=1.0\columnwidth]{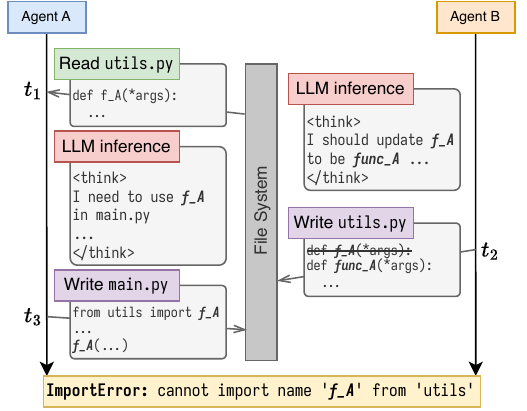}
\vspace{0.2cm}
\caption{\textbf{Stale read hazard in multi-agent coding.} Agent A reads \texttt{utils.py} and enters a long inference phase while implementing \texttt{main.py}. Concurrently, Agent B refactors \texttt{utils.py}, renaming \texttt{f\_A} into \texttt{func\_A}. Both agents act correctly in isolation, yet the interleaving yields a broken import, a classic concurrency anomaly amplified by long LLM inference windows.}
\label{fig:stale-read-mas}
\end{figure}

\textbf{In this position paper, we argue that many communication and coordination failures in MAS are fundamentally concurrency control problems, which should be considered as a prioritized bottleneck for building efficient, scalable multi-agent systems.}

Our claim targets MAS in which agents read or modify shared mutable state during long inference windows, including shared repositories, blackboard memories, message buffers, and embodied world states.
Systems with disjoint inputs face lower concurrency risk and may be bottlenecked by reasoning, planning, or communication quality.
This scope complements MAS surveys that organize workflows, infrastructure, and collaboration patterns \citep{chen2024survey,li2024survey,tran2025multiagent,guo2024large}; our contribution is the concurrency-control lens that connects independently developed coordination mechanisms.

\begin{table}[htb]
\centering
\caption{Recent evidence that concurrency mechanisms affect MAS outcomes.}
\label{tab:evidence}
\fontsize{9pt}{12pt}\selectfont
\begin{tabularx}{\columnwidth}{@{}p{1.4cm}YY@{}}
\toprule
\textbf{Source} & \textbf{Concurrency signal} & \textbf{Reported effect} \\
\midrule
CAID & worktree isolation, merge validation & 63.3\% isolated vs.\ 55.5\% unisolated; single-agent 57.2\% \\
CodeR & dependency scheduling & 22\% vs.\ 10\% resolved after removing the task graph \\
Silo-Bench & barriers, state conflicts & 67.1\% of failures; RCC reaches 100\% at high contention \\
MegaAgent & parallel scheduling & 800s vs.\ 4505s without parallel group execution \\
SagaLLM & saga transactions & correct reactive planning where baseline LLM planners fail \\
\bottomrule
\end{tabularx}
\end{table}

\begin{table}[htb]
\centering
\caption{Concurrency-attributable failures are a substantial fraction.}
\label{tab:fraction}
\fontsize{9pt}{12pt}\selectfont
\begin{tabularx}{\columnwidth}{@{}p{2cm}YYp{2cm}@{}}
\toprule
\textbf{Failure Mode} & \textbf{Source} & \textbf{Rate} & \textbf{Concurrency Root} \\
\midrule
Premature submission & Silo-Bench & 37.2\% & Missing sync barriers \\
Consensus failure & Silo-Bench & 29.9\% &  Concurrent conflicting states \\
Inter-agent misalignment & MAST & 36.9\% &  Stale reads, inconsistent state \\
Coordination overhead & Silo-Bench & RCC $\leq$ 100\% & 	Concurrency scaling penalty \\
\bottomrule
\end{tabularx}
\end{table}

To illustrate this perspective concretely, consider a motivating example involving two coding agents collaboratively developing software on a shared file system. 
Agent $A$ reads an existing utility module \texttt{utils.py} and subsequently implements a new feature in \texttt{main.py} that imports and invokes a function \texttt{f\_A} from the utility module. 
Concurrently, Agent $B$ is refactoring \texttt{utils.py}, renaming \texttt{f\_A} to \texttt{func\_A}. 
Unfortunately, between the moment $A$ reads the utility module and the moment $A$ completes its implementation, $B$ writes the refactored version. 
The result: both agents complete their individual tasks correctly from their own perspectives, yet the system enters an inconsistent state where \texttt{main.py} contains broken imports. 
This failure superficially appears to be a coordination or communication problem. 
However, examining it through the lens of concurrent systems reveals a classic concurrency hazard: a \emph{stale read} leading to an inconsistent state.

What makes such hazards particularly prevalent in MAS is a fundamental \emph{temporal asymmetry}: LLM inference time (e.g., the duration of the ``thinking'' phase) is typically orders of magnitude longer than the execution time of tool actions, dramatically expanding the window during which interleaved operations can produce conflicts. While an agent reasons for seconds or even minutes, other agents may modify the shared environment multiple times, invalidating the assumptions underlying the first agent's decisions.

The remainder of this paper is organized as follows.
We first demonstrate that common MAS failure modes can be systematically understood as \textbf{concurrency hazards} (\S\ref{sec:hazards}), revealing that diverse ``coordination problems'' share a common structure rooted in concurrent access to shared state.
We then present \textbf{recommendations} for concurrency control in MAS design (\S\ref{sec:recommendations}, Table~\ref{tab:mas-cc-design}), addressing objectives, system-level mechanisms, and trade-offs among correctness, efficiency, and scalability.
Finally, we discuss open challenges and issue a \textbf{call to action} (\S\ref{sec:conclusion}) for cross-disciplinary collaboration between ML and systems-oriented communities.

\section{Concurrency Hazards in Disguise}
\label{sec:hazards}

\begin{table*}[t]
\centering
\caption{Design space for MAS concurrency control. Trade-offs evaluated against: task success (\textbf{S}), compatibility (\textbf{C}), efficiency (\textbf{E}), inference cost (\textbf{I}). Arrows: $\uparrow$ improves, $\downarrow$ degrades.}
\label{tab:mas-cc-design}
\fontsize{8pt}{11pt}\selectfont
\begin{tabular}{@{}clp{12cm}@{}}
\toprule
\textbf{Layer} & \textbf{Decision} & \textbf{Options (Trade-offs)} \\
\midrule
\multirow{8}{*}{\rotatebox[origin=c]{90}{\small \centering \textit{System Design}}} 
& Isolation level & Weak/Read Committed (E$\uparrow$, S$\downarrow$: more parallelism, risks anomalies) $\leftrightarrow$ Strong/Serializable (S$\uparrow$, E$\downarrow$) \\
& Control strategy & Pessimistic/locking (S$\uparrow$, E$\downarrow$: blocks during long inference) vs.\ Optimistic/validation (E$\uparrow$, I$\downarrow$: wastes compute on abort) \\
& Versioning & Single-version (simple) vs.\ MVCC (E$\uparrow$: readers never block writers; C$\downarrow$: added complexity) \\
& Transaction granularity & Fine-grained/single action (E$\uparrow$, I$\downarrow$: shorter conflicts, higher overhead) vs.\ Coarse/subtask (I$\uparrow$, S$\downarrow$: expensive rollbacks) \\
& Transaction boundaries & Explicit \texttt{BEGIN/COMMIT} (C$\uparrow$, I$\downarrow$: flexible, requires model understanding) vs.\ Implicit/system-inferred (I$\uparrow$, C$\downarrow$) \\
& Lock/resource granularity & Coarse/files (I$\uparrow$, E$\downarrow$) vs.\ Fine/functions (E$\uparrow$, I$\downarrow$: more parallelism, more metadata) \\
\midrule
\multirow{6}{*}{\rotatebox[origin=c]{90}{\small \centering \textit{Infrastructure}}}
& Backend system & Custom (C$\uparrow$: tailored semantics) vs.\ Existing DB/Git/FS (S$\uparrow$, C$\uparrow$: mature guarantees, may not fit agent semantics) \\
& Version control integration & Branch-per-subtask (S$\uparrow$: isolation; E$\downarrow$: merge overhead) vs.\ Validation-at-merge (E$\uparrow$, S$\downarrow$: deferred conflict detection) \\
& Inference optimization & Standard vs.\ Optimized batching/speculation/quantization (E$\uparrow$, S$\uparrow$: shorter transactions reduce conflict window) \\
& Checkpointing & None (simple) vs.\ KV-cache checkpointing (I$\uparrow$: efficient rollback without full recomputation; C$\downarrow$: engine support required) \\
\midrule
\multirow{4}{*}{\rotatebox[origin=c]{90}{\small \centering \textit{Model}}} 
& Concurrency training & None vs.\ SFT/RL on conflict scenarios (S$\uparrow$: better anticipation/resolution; I$\downarrow$: requires data and compute) \\
& Prompt intervention & Generic vs.\ Concurrency-aware prompts (I$\uparrow$: low cost; S$\pm$: limited, brittle guarantees) \\
& Task decomposition & Overlapping resources (E$\uparrow$, S$\downarrow$) vs.\ Disjoint partitioning (S$\uparrow$, C$\downarrow$: requires upfront design effort) \\
& Failure feedback & Opaque ``retry'' (I$\uparrow$: simple) vs.\ Semantic conflict details (S$\uparrow$, I$\downarrow$: enables adaptation, requires model capability) \\
\bottomrule
\end{tabular}
\end{table*}

Many failures in multi-agent systems, often framed as coordination or communication issues, can be understood as classic concurrency hazards. We present representative failure cases and map them to established concurrency control concepts, showing that these challenges mirror problems long studied in systems and database research~\citep{bernstein1981concurrency,bernstein1987concurrency,weikum2001transactional}. This perspective motivates treating concurrency control as a core design principle for MAS and grounding coordination heuristics in explicit concurrency semantics.

\subsection{Failure Examples}
\label{subsec:failure-examples}

We present four representative failure scenarios in multi-agent systems that are often attributed to poor coordination or communication, but are more precisely explained as classic concurrency hazards over shared state.

\noindent\textbf{Stale Read.}
In a collaborative coding MAS~\citep{qian2024chatdev,hong2024metagpt}, Agent $A$ reads a configuration file \texttt{config.yaml} and implements a client module based on the observed endpoints. Concurrently, Agent $B$ updates the configuration to reflect new infrastructure. When $A$ later writes its code, it relies on outdated assumptions, resulting in an inconsistent system state despite both agents acting correctly in isolation. This is a standard stale-read anomaly.

\noindent\textbf{Lost Update.}
Two agents independently modify different parts of a shared file \texttt{utils.py}. Agent $A$ optimizes one function, while Agent $B$ fixes a bug in another. Because both read the same initial version and write back independently, the later write overwrites the earlier one, silently discarding one agent’s contribution. This lost update occurs without either agent detecting a conflict.

\noindent\textbf{Stale Correction.}
In message-based MAS~\citep{li2023camel,wu2024autogen}, Agent $A$ broadcasts a plan, Agent $B$ sends a correction, but Agent $C$ begins executing the original plan before receiving the update. Although the correction is sent, it is not applied atomically with respect to all agents, leading to incorrect execution driven by outdated information.

\noindent\textbf{Action-Message Desynchronization.}
In embodied environments such as Minecraft-based benchmarks~\citep{fan2022minedojo,wang2023voyager,dong2024villageragent}, agents coordinate via both messages and world-altering actions. An agent may act on a message describing an intended state that has already been invalidated by another agent’s concurrent action. As a result, agent beliefs and the true environment state diverge, even when communication is logically consistent.

\noindent\textbf{Implicit Assumptions in Existing Systems.}
These hazards are present in current MAS architectures. Systems such as MAGIS~\citep{tao2024magis} improve reliability through structured orchestration and post-hoc Git-based merging, but implicitly assume benign interleavings during concurrent reasoning. Conflicts are detected only after agents have already performed expensive inference on incompatible assumptions. This reflects a broader pattern across MAS frameworks that rely on orchestration while leaving concurrency semantics underspecified.
Recent systems show the same pattern from another angle. CAID uses isolated worktrees and merge-time validation, CodeR uses a task graph to order dependent work, SagaLLM uses saga-style compensation, and MegaAgent emphasizes parallel scheduling~\citep{geng2026effective,chen2024coder,chang2025sagallm,wang2024megaagent}. These mechanisms correspond to optimistic isolation, dependency scheduling, transactional recovery, and concurrent scheduling, even when they are introduced without explicit concurrency-control terminology.

\subsection{A Unified Framework}
\label{subsec:framework}

The failure examples above, though arising in different application contexts, share a common structure. We now develop a unified framework that reveals these failures as violations of fundamental concurrency properties, drawing on classical concurrency control theory \citep{bernstein1987concurrency,weikum2001transactional,adya1999weak}.

\noindent\textbf{Formalizing Multi-Agent Systems.}
We model a multi-agent system as a collection of $n$ agents with respective internal states $\{\mathcal{A}^1, \mathcal{A}^2, \ldots, \mathcal{A}^n\}$, all operating within a \emph{shared} environment $\mathcal{E}$. Each agent $j$ executes a sequence of actions $a^j_1, a^j_2, \ldots$ according to its own local ordering, but these sequences interleave concurrently across agents, i.e., actions may overlap temporally, and conflicting operations on shared resources can occur.

We classify actions based on whether they modify the shared environment. An action is \emph{side-effect-free} (a \emph{read}) if it leaves $\mathcal{E}$ unchanged; examples include reading a file, observing game state, or retrieving messages. An action is \emph{side-effecting} (a \emph{write}) if it modifies $\mathcal{E}$; examples include writing a file, placing a game object, or broadcasting a message.

We further distinguish between \emph{agent-local state} and \emph{environment state} based on \emph{mutability by other agents}: environment state can be modified by any agent's writes, while agent-local state (e.g., context window, reasoning trace) can only be directly modified by that agent itself. In coding MAS, the shared file system constitutes environment state. In message-passing MAS, each agent's mailbox, which is writable by other agents via message sends, is environment state. In game environments, both world state and communication channels are environment state.

A key observation is that LLM-based reasoning is side-effect-free with respect to $\mathcal{E}$, but requires substantial time to complete. This \emph{temporal asymmetry}, i.e., inference spans seconds to minutes while tool execution completes in milliseconds, dramatically expands the window during which interleaved operations can produce conflicts.

\noindent\textbf{Consistency and Isolation.}
We identify two properties whose violation underlies MAS failures, drawing on the ACID properties from database theory \citep{haerder1983principles}:

\emph{Consistency} requires that concurrent operations do not leave the system in a conflicting or contradictory state. In MAS, this means the combined effects of all agents' actions should be mutually compatible: two agents should not simultaneously assume exclusive use of the same resource, and actions taken based on observations should remain valid when executed. Consistency is a \emph{global} property concerning system-wide coherence.

\emph{Isolation} requires that each agent's execution proceeds ``as if'' it were the only agent in the system. 
Formally, the outcome should be equivalent to some serial execution of all agents' actions \citep{weikum2001transactional}. Isolation is an \emph{agent-centric} property: no agent should observe another agent's incomplete operation sequence. 
The classical notion of \emph{serializability}, that concurrent execution should be equivalent to some sequential ordering, captures the strongest form of isolation \citep{papadimitriou1979serializability}.

These properties are complementary: consistency ensures global coherence, while isolation ensures local predictability. Violations of either lead to the anomalies described above.

\noindent\textbf{Mapping Failures to the Framework.}
In \emph{coding MAS}, the \emph{write-after-read} failure violates both consistency (Agent $A$'s implementation conflicts with Agent $B$'s configuration changes) and isolation (Agent $A$'s read was invalidated by Agent $B$'s concurrent write). The \emph{lost update} failure similarly violates both: consistency is violated because only one agent's modifications persist, and isolation is violated because one agent's write was silently overwritten.

In \emph{message-based MAS}, the \emph{stale correction} failure maps directly to the write-after-read pattern: Agent $C$ reads the original plan, Agent $B$ writes the correction, but Agent $C$'s actions proceed based on the stale read. In \emph{cooperative game MAS}, the \emph{action-message desynchronization} failure violates consistency because the coordination message and world state become inconsistent, and isolation because Agent $B$'s operation is disrupted by Agent $C$'s uncoordinated action.

\noindent\textbf{Implications for MAS Design.}
The unified framework identifies failures documented across diverse MAS applications as systematic consequences of uncontrolled concurrency. The database and distributed systems communities have developed extensive techniques to address precisely these problems: locking protocols \citep{eswaran1976notions}, optimistic concurrency control \citep{kung1981optimistic}, multi-version concurrency control \citep{reed1978naming}, and various isolation levels \citep{berenson1995critique,adya1999weak}. For MAS research, the central task is adapting classical concurrency control mechanisms to the unique characteristics of LLM-based agents: long inference latencies, natural-language interfaces, and imperfect adherence to specifications. We address this task in the following section.

\section{Concurrency Control for MAS}
\label{sec:recommendations}

Having established that many MAS failures stem from concurrency hazards, we now address the central question: how should MAS designers approach concurrency control? 
This section articulates design objectives and examines the mechanisms available to system architects, organized across three complementary layers (Table~\ref{tab:mas-cc-design}): system design, infrastructure and model capabilities. Together, these layers provide a comprehensive framework for understanding where concurrency control responsibilities reside and how they may be systematically addressed.

\subsection{Design Objectives}
\label{subsec:objectives}

Effective concurrency control for MAS should balance multiple, often competing objectives. We identify four primary dimensions that collectively define the quality of a concurrency control solution.

\noindent\textbf{Task Success Rate.}
The fundamental objective is ensuring that concurrent agent execution produces correct results. What constitutes ``success'' depends on the application: in coding tasks, it may mean passing all tests without introducing bugs; in collaborative writing, it may mean producing coherent output without contradictory edits. A concurrency control mechanism should maximize task success rate across diverse MAS workloads. Existing MAS benchmarks~\citep{liu2024agentbench,cemri2025multi,jimenez2024swe,zhou2024webarena} provide starting points for measuring success, though they largely ignore concurrency-specific failure modes.

\noindent\textbf{Compatibility.}
Concurrency control mechanisms should accommodate diverse MAS architectures: centralized orchestrators~\citep{wu2024autogen}, decentralized peer-to-peer designs~\citep{li2023camel}, hybrid topologies~\citep{hong2024metagpt}, and hierarchical structures~\citep{qian2024chatdev}. A mechanism tightly coupled to a specific communication pattern or agent architecture limits reusability. Ideally, concurrency control primitives should be orthogonal to the MAS coordination strategy, enabling modular integration without redesigning the entire system.

\noindent\textbf{Execution Efficiency.}
Concurrency control inherently restricts parallelism to ensure correctness, creating tension with performance goals. Key efficiency metrics include: (i) \emph{end-to-end task completion time}: the wall-clock duration for the MAS to finish a task; (ii) \emph{effective parallelism}: the degree to which agents operate concurrently without blocking; and (iii) \emph{abort/retry rate}: the frequency of transaction rollbacks due to detected conflicts. An ideal mechanism minimizes completion time while maintaining correctness guarantees.

\noindent\textbf{Inference Cost.}
MAS concurrency control incurs costs beyond those of traditional distributed systems: direct overhead from tool calls for lock acquisition or conflict detection; model comprehension cost, as LLMs may struggle with concurrency primitives and require extended prompts or fine-tuning; and context window consumption from concurrency-related instructions and state tracking. Critically, mechanisms requiring agents to explicitly manage locks or reason about conflicts demand far greater model sophistication than transparent system-level enforcement.

\subsection{System Design for MAS}
\label{subsec:system}

We now examine the theoretical foundations and concrete design choices for MAS concurrency control, spanning isolation levels, control strategies, primitives, and integration with broader MAS mechanisms.

\subsubsection{Theoretical Foundations}

\noindent\textbf{Isolation Levels.}
Database systems formalize correctness through \emph{isolation levels}, which specify the degree to which transactions are protected from interference~\citep{berenson1995critique,adya1999weak}. The classical hierarchy includes, from weakest to strongest: Read Uncommitted, Read Committed (RC), Repeatable Read (RR), Snapshot Isolation (SI), Serializable (SER), and Strict Serializable (SSER)~\citep{bernstein1987concurrency,fekete2005making,papadimitriou1979serializability}. Each level prevents progressively more anomalies (e.g., dirty reads, non-repeatable reads, phantom reads, and write skew) but typically incurs higher overhead~\citep{cahill2009serializable}.

For MAS, the appropriate isolation level depends critically on the application domain. A multi-agent coding system modifying a shared codebase may require serializability to prevent silent corruption from conflicting edits. In contrast, a multi-agent debate system where agents operate on largely disjoint logical resources (their own arguments) may tolerate weaker guarantees like Read Committed, trading strict isolation for higher throughput. The general idea is that stronger isolation provides better guarantees to agents but may reduce performance and increase implementation complexity, which is a tradeoff that should be calibrated per application.

\noindent\textbf{Pessimistic vs.\ Optimistic Control.}
Concurrency control strategies fall into two broad categories~\citep{bernstein1981concurrency,kung1981optimistic}. \emph{Pessimistic} approaches, such as two-phase locking (2PL), acquire locks before accessing resources, blocking other agents until locks are released~\citep{eswaran1976notions,gray1976granularity}. This prevents conflicts proactively but may cause deadlocks (circular waits) and reduces parallelism. \emph{Optimistic} approaches allow agents to proceed without locks, validating at commit time that no conflicts occurred; conflicting transactions abort and retry. Optimistic methods suit low-contention workloads but may thrash under high contention.

A crucial distinction from traditional systems is the \emph{temporal asymmetry} in MAS: LLM inference typically spans seconds to minutes, while simple tool calls complete in milliseconds. This asymmetry fundamentally alters the calculus. Under pessimistic control, a lock held during LLM reasoning blocks other agents for extended periods, severely degrading parallelism. Under optimistic control, aborts waste substantial compute: an agent may reason for minutes only to have its transaction invalidated. Neither approach dominates; the choice depends on contention levels, task structure, and the relative costs of blocking versus retry.
Recent coding-agent evidence illustrates this trade-off: CAID reports that isolated branches with merge-time validation outperform a shared-workspace multi-agent baseline, suggesting that optimistic isolation is effective when conflicts can be validated cheaply at integration time~\citep{geng2026effective}.

\noindent\textbf{Multiversion Concurrency Control.}
Multiversion concurrency control (MVCC) offers a middle ground by maintaining multiple versions of data~\citep{reed1978naming,bernstein1987concurrency}. Readers access consistent snapshots without blocking writers, enabling the principle that ``readers never block writers and writers never block readers.'' MVCC naturally supports Snapshot Isolation~\citep{fekete2005making} and forms the basis for concurrency control in many modern databases. For MAS, MVCC-style approaches could allow agents to observe consistent environment states while others make modifications, with conflict detection deferred to commit time. The challenge lies in defining ``versions'' for arbitrary agent environments beyond traditional databases.

\subsubsection{Primitive Design}

\noindent\textbf{Transaction Granularity.}
A core design decision is the unit of atomicity. Transactions may range from individual actions (e.g., a single tool call) to entire agent subtasks (e.g., ``implement feature X''). Fine-grained transactions shorten conflict windows but incur higher coordination overhead, while coarse-grained transactions reduce overhead at the cost of higher conflict probability and more expensive rollbacks.

Another question is whether transaction boundaries are \emph{explicit}, via primitives such as \texttt{BEGIN}/\texttt{COMMIT}, or \emph{implicit}, inferred by the system from agent behavior or task structure. Explicit control gives agents flexibility but assumes that LLMs can reliably reason about transactional semantics. Implicit control lowers cognitive burden but risks poorly chosen boundaries that either fragment logical work or bundle too much state into a single transaction.

\noindent\textbf{Resource Abstraction and Lock Types.}
Lock-based designs should define what constitutes a lockable resource: entire files, individual functions, database rows, or higher-level semantic objects (e.g., ``a calendar slot''). A \emph{fine-grained resource} is the smallest unit to which access control applies. In a codebase, file-level locking treats \texttt{utils.py} as one resource, while function-level locking treats \texttt{parse\_config} and \texttt{validate\_input} as distinct resources. Function-level locking allows two agents to modify different functions concurrently, whereas file-level locking serializes them unnecessarily. This mirrors row-level and table-level locking in databases~\citep{gray1976granularity}. Finer-grained resources enable more parallelism but require more complex tracking and metadata.

Lock modes further regulate access~\citep{gray1976granularity}. \emph{Shared (read) locks} permit concurrent observation, while \emph{exclusive (write) locks} serialize modifications. This distinction aligns well with MAS workloads, where reads are frequent and typically safe to overlap, whereas writes should be carefully coordinated. More expressive mechanisms, such as intention or predicate locks, may be useful for hierarchical or semantic resources.

\noindent\textbf{Conflict Detection and Rollback.}
Optimistic control requires detecting conflicts by tracking \emph{read sets} and \emph{write sets}, then validating that no incompatible updates occurred concurrently~\citep{kung1981optimistic,adya1999weak}. Software transactional memory (STM) offers a reference model for such tracking~\citep{shavit1995software,harris2005composable}, though MAS environments are more heterogeneous than the shared-memory settings STM targets.
This heterogeneity is an analytical distinction: MAS resources extend beyond memory locations to files, prompts, tool outputs, mailboxes, external APIs, and natural-language commitments.
Beyond conflict detection within the MAS runtime itself, recent black-box and history-based database isolation checkers suggest a complementary direction: treating MAS executions as observable histories and using an external monitor to check whether interactions satisfy different consistency or isolation levels~\citep{cobra20,plume24,verystrong25}.

When conflicts arise, aborted transactions should be rolled back. In MAS, rollback is challenging because agent state includes external writes, internal context, and intermediate reasoning. Techniques such as KV-cache checkpointing~\citep{kwon2023efficient} may allow partial rewinding without full re-inference. However, actions with irreversible side effects (e.g., sending emails or invoking external APIs) fundamentally limit rollback and should be treated with special care.

\subsubsection{MAS Integration}

\noindent\textbf{Synchronization Granularity.}
MAS frameworks should decide on the \emph{minimal synchronization unit}: the finest granularity at which agents coordinate. One approach uses \emph{round-based synchronization}: agents act in discrete rounds, with a barrier ensuring all agents complete round $k$ before any begins round $k+1$. This simplifies reasoning about concurrent state but limits parallelism and may be inefficient when agents' tasks have heterogeneous durations. The alternative is \emph{fully asynchronous} execution where agents proceed independently, relying solely on concurrency control primitives (locks, validation) for coordination. This maximizes parallelism but requires more sophisticated conflict handling.

The synchronization granularity should align with transaction boundaries. If transactions span multiple rounds in a round-based system, mid-transaction preemption creates complex partial states. Conversely, if transactions complete within single rounds, the round barrier naturally provides commit points.

\noindent\textbf{Inference Interruption.}
A distinctive aspect of MAS is whether ongoing inference can be interrupted. An agent mid-reasoning may be holding logical locks or accumulating reads that will be validated at commit. If another agent's commit invalidates these reads, should the first agent be \emph{preempted} (immediately aborted) or allowed to complete then fail validation? Preemption avoids wasted compute but requires mechanisms to safely interrupt LLM inference and restore state. Allowing completion then validation is simpler but wastes resources when conflicts are predictable.

Related questions include whether agents can \emph{voluntarily} yield execution (cooperative scheduling) and how interrupted agents resume from the beginning of the transaction, from a checkpoint, or with partial context preserved via KV cache.

\noindent\textbf{Semantic Feedback on Failures.}
When a transaction fails due to conflict detection, lock timeout, or validation failure, the agent receives an error signal. A key design choice is whether to provide \emph{semantic explanations} of failures. A rich explanation might state: ``your write to \texttt{utils.py} conflicted with Agent B's concurrent modification of the \texttt{parse\_config} function.'' An opaque signal simply indicates ``transaction aborted, please retry.''

Semantic feedback enables agents to reason about conflicts and adapt strategies---perhaps avoiding the contested resource, coordinating with the conflicting agent, or decomposing the task differently. However, this requires models capable of interpreting and acting on such information, a capability that may require specific training or careful prompting. Opaque signals are simpler but provide no guidance for conflict resolution.

\noindent\textbf{Integration with Existing Pipelines.}
A practical question is how to augment existing MAS workflows, which typically assume benign interleavings, with explicit concurrency semantics. Consider systems like MAGIS~\citep{tao2024magis} that use structured role specialization and Git-based collaboration. Several ``drop-in'' strategies could add concurrency guarantees without redesigning the entire pipeline:

\begin{itemize}[nosep,noitemsep,topsep=0pt,leftmargin=*,wide=0pt]
    \item \emph{Branch-per-subtask}: Each agent subtask operates on an isolated Git branch, with merges gated by automated conflict detection and semantic validation (e.g., test execution, type checking).
    \item \emph{Transactional file edits}: Wrap multi-file modifications in lightweight transactions that abort if concurrent changes to the same files are detected before commit.
    \item \emph{Validation-at-merge}: Defer validation to merge time and enrich merge checks with semantic constraints beyond textual diff resolution.
\end{itemize}

These strategies trade off eager conflict prevention against deferred conflict detection. Branch-per-subtask maximizes isolation but complicates cross-branch dependencies; validation-at-merge preserves existing workflows but may waste computation on branches that ultimately conflict.

\subsection{Infrastructure Support}
\label{subsec:infrastructure}

MAS designers can leverage existing infrastructure that already provides well-tested concurrency primitives.

\noindent\textbf{File Systems and Databases.}
File systems offer atomic operations and advisory locking, while relational databases provide full ACID transactions with configurable isolation levels~\citep{haerder1983principles,weikum2001transactional}. Vector databases increasingly expose transactional semantics as well~\citep{pinecone2023,weaviate2023}. Structuring agent interactions around such systems allows MAS to inherit mature concurrency guarantees and performance optimizations, offloading much of the correctness burden.

\noindent\textbf{Version Control Systems.}
Version control systems such as Git support concurrent work via branching and explicit merge-based conflict resolution. Recent work explores Git as coordination infrastructure for MAS. EvoGit~\citep{huang2025evogit} uses Git's DAG structure to enable asynchronous multi-agent development, isolating agents on branches and synchronizing via merges while avoiding global serialization. This design favors scalability and fault isolation, but defers semantic conflict resolution to merge time, where textual diffs may not capture program-level invariants.

\noindent\textbf{Context Management Infrastructure.}
Git-like abstractions have also been applied to agent context management. GCC~\citep{wu2025git} provides branching and merging for agent memory, enabling efficient context reuse and performance gains. However, it targets primarily single-agent settings and does not define concurrency semantics for multi-agent interleavings over shared context.

Across these systems, a common pattern emerges: infrastructure supplies powerful \emph{mechanisms} (transactions, branches, merges), while higher-level frameworks must still define \emph{policies}: when to isolate, what conflicts are acceptable, and how to enforce semantic invariants. Bridging this gap between infrastructure capabilities and concurrency semantics is central to the agenda we advocate.

\noindent\textbf{Inference Engine Efficiency.}
The temporal asymmetry that exacerbates MAS concurrency issues (i.e., long inference latencies relative to tool execution) can be mitigated by faster inference. Advances including continuous batching~\citep{yu2022orca}, paged attention~\citep{kwon2023efficient}, speculative decoding~\citep{leviathan2023fast}, and quantization~\citep{frantar2022gptq} reduce per-token latency and thus shrink the window during which an agent holds logical locks or accumulates stale reads. From a concurrency control perspective, faster inference is equivalent to shorter transactions, directly reducing conflict probability.

\noindent\textbf{Efficient Checkpointing for Rollback.}
Optimistic concurrency control requires efficient rollback. In LLM inference, this translates to restoring the KV cache to a prior state. Modern inference engines with paged KV cache management~\citep{kwon2023efficient} can potentially support \emph{forking} (copying cache state for speculative execution) and \emph{rollback} (discarding cache entries added after a checkpoint). Explicit support for these operations would enable efficient transaction abort without full recomputation. Cache sharing across agents with common prefixes~\citep{zheng2024sglang} further reduces the cost of retry by amortizing prompt processing.

\subsection{Model Capabilities}
\label{subsec:model}

Complementing system-level approaches, the underlying LLMs can be improved to exhibit concurrency-aware behavior through evaluation, training, and inference-time interventions.

\subsubsection{Benchmarking Concurrency Awareness}

A prerequisite for improving model capabilities is measuring them. We advocate for benchmarks that evaluate: (i) an agent's ability to \emph{anticipate} conflicts given knowledge of other agents' goals and current activities; (ii) an agent's ability to \emph{detect} conflicts from environmental feedback (e.g., failed writes, stale reads); and (iii) an agent's ability to \emph{resolve} conflicts through negotiation, backing off, or requesting arbitration. Such benchmarks could extend existing MAS evaluation suites~\citep{liu2024agentbench,cemri2025multi} with concurrency-specific metrics: conflict rate, resolution success rate, wasted computation from aborts, and throughput under contention.

A particularly valuable benchmark type would measure agent behavior \emph{without} explicit concurrency control, revealing how naturally (or poorly) current models handle concurrent settings and what failure modes emerge most frequently.

\subsubsection{Training for Concurrency Awareness}

Given appropriate benchmarks and environments, LLMs can be trained to exhibit concurrency-aware behavior. Supervised fine-tuning (SFT) on traces of successful conflict anticipation, detection, and resolution can instill basic patterns. Reinforcement learning (RL) in multi-agent environments, building on the substantial MARL literature~\citep{lowe2017multi,foerster2018learning,yu2022surprising,zhang2021multi}, can train agents to coordinate under contention, learning when to yield, retry, or escalate.

The reward signal should reflect both task success and coordination efficiency, penalizing unnecessary conflicts and wasted retries. Centralized training with decentralized execution can mitigate non-stationarity when agents share rewards and training infrastructure. The harder case is heterogeneous deployment, where agents from different providers or training regimes coordinate without prior joint training, a setting closely related to Ad Hoc Teamwork~\citep{stone2010adhoc,barrett2015cooperating}. In such deployments, as one agent's policy or system prompt changes, the effective environment for other agents shifts, complicating convergence and reliable coordination~\citep{hernandez2017survey}. Techniques from cooperative MARL, including centralized training with decentralized execution and opponent modeling, may help address this challenge.

\subsubsection{Inference-Time Interventions}

Without modifying models, some concurrency-aware behavior can be induced through prompt design and task structuring.

\noindent\textbf{Prompt Engineering.}
Agents can be explicitly informed that they operate in a concurrent environment with shared resources and guided toward conservative behaviors, such as checking for concurrent edits, acquiring resources in a consistent order, or retrying after failed writes. Such prompts rely on the base model's instruction-following ability and whatever concurrency knowledge it has implicitly acquired during pretraining, and thus offer limited and brittle guarantees.

\noindent\textbf{Task Decomposition and Orchestration.}
Contention can often be reduced by decomposing tasks so that agents operate on largely disjoint resources, with hierarchical approaches dividing responsibilities between strategic and tactical agents to reduce coordination complexity. While perfect partitioning is rarely possible, careful decomposition can substantially lower conflict frequency. These approaches trade upfront design effort for reduced runtime conflicts, but scale poorly to novel or highly interdependent workloads where explicit concurrency control becomes necessary.

\section{Alternative Views}
\label{sec:alternatives}

\noindent\textbf{Foundation model capabilities alone suffice.}
One view holds that sufficiently capable models can resolve conflicts without explicit concurrency control. 
Conversation-centric frameworks such as AutoGen exemplify the appeal of relying on model capability and dialogue structure for coordination~\citep{wu2024autogen}.
However, as agent count grows, conflict probability approaches certainty regardless of model quality. 
Natural-language negotiation wastes inference budget, coordination embedded in model weights is neither debuggable nor auditable, and robust concurrency awareness would require training data that does not exist. 
Model behavior also remains non-deterministic in practice~\citep{atil2024non,kahng2024llm}.

\noindent\textbf{Traditional systems techniques suffice.}
Another view argues that existing database transactions and locks can be directly applied to MAS. However, LLM agents are semantic-driven, non-deterministic~\citep{atil2024non}, and incur inference costs orders of magnitude larger than system-level operations. These mismatches make direct transplantation ineffective and instead call for system--model co-design.

\noindent\textbf{Eventual consistency and convergence suffice.}
A third view suggests that eventual consistency (e.g., via CRDTs) is sufficient~\citep{shapiro2011conflict,kleppmann2017json}. This conflates convergence with semantic correctness. While replicas may eventually agree, application invariants can be violated when agents reason over long inference windows based on invalidated state. CRDTs suit independent operations with well-defined merge semantics, but MAS often rely on cross-resource invariants that should remain stable during reasoning. The key question is what guarantees agents can rely on while they think and act.

\noindent\textbf{Existing community tools already cover the issue.}
Existing tool (SE branching and CI, MARL coordination, and distributed consistency protocol) are valuable starting points, but LLM-MAS uniquely combines shared-state hazards with long, costly, language-mediated inference. The open problem is exposing these as agent-facing primitives with clear isolation, validation, and rollback semantics.

\section{Call to Action}
\label{sec:conclusion}

Many coordination failures in multi-agent systems are, at their core, concurrency control problems. The anomalies well-known from parallel programs and distributed databases resurface when agents share mutable state, amplified by long LLM inference windows and non-determinism.
Closing this gap requires cross-disciplinary collaboration. ML researchers should treat concurrency as a first-class systems concern rather than emergent behavior to be prompted around, while researchers in PL and DB should adapt classical techniques to LLM-based agents through careful redesign. Neither perspective alone is sufficient.

\subsection{For ML Researchers}

\noindent\textbf{Recognize concurrency as a distinct failure mode.}
When MAS exhibit timing-dependent errors or coordination breakdowns despite locally correct actions, the root cause is often concurrent access to shared state under long inference windows.
Treating these as generic coordination problems obscures the underlying issue.
Explicit recognition is the first step toward systematic solutions.

\noindent\textbf{Design benchmarks with contention in mind.}
Current evaluations~\citep{liu2024agentbench,jimenez2024swe,zhou2024webarena} report aggregate success rates but cannot distinguish capability failures from concurrency failures.
A system achieving 70\% success may fail 20\% of the time due to stale reads alone, yet this remains invisible in current metrics.
New benchmarks should vary contention levels systematically and measure conflict frequency, resolution success, wasted computation from aborts, and effective parallelism under contention.

\noindent\textbf{Train for concurrency awareness.}
Pretraining corpora contain extensive coverage of sequential programming but minimal exposure to concurrent coordination patterns such as lock denial, abort recovery, or conflict anticipation under partial observability.
Whether models can acquire robust concurrency-aware behavior via prompting alone or require explicit training with multi-agent RL~\citep{lowe2017multi,foerster2018learning} remains an open and consequential question.

\subsection{For Researchers in Systems-Oriented Fields}

\noindent\textbf{Adapt, not transplant.}
Databases and distributed systems offer mature foundations: serializability theory~\citep{papadimitriou1979serializability}, isolation hierarchies~\citep{berenson1995critique,adya1999weak}, and protocols such as 2PL~\citep{eswaran1976notions}, OCC~\citep{kung1981optimistic}, MVCC~\citep{reed1978naming}, and SSI~\citep{cahill2009serializable}.
However, LLM agents interact through natural language and are non-deterministic, making direct transplantation ineffective.
The core challenge is determining which abstractions transfer, which require redesign, and what new primitives are needed.

\noindent\textbf{Design for latency asymmetry.}
Database transactions complete in microseconds; LLM inference takes seconds to minutes.
Pessimistic locking blocks unacceptably at these timescales; optimistic control wastes expensive computation on aborts.
New protocols should explicitly account for this asymmetry, potentially via interruptible inference, KV-cache rollback, or contention-aware scheduling.
The success criterion is preserving task success while improving effective parallelism and reducing wasted tokens under controlled contention.

\noindent\textbf{Build agent-friendly concurrency infrastructure.}
MAS currently lack the primitives that make concurrency manageable in traditional systems: versioning, conflict detection, and rollback semantics exposed through interfaces that agents can reliably use.
Equally important are observability and debugging tools that make concurrency failures explicit and actionable.
Without such tooling, diagnosing whether a failure originates from a race condition, a coordination misalignment, or a model-level error remains prohibitively difficult in practice.
Building this infrastructure is a prerequisite for principled progress.

\section*{Acknowledgements}
We thank Dr. Si Liu at ETH Zürich for constructive discussions on concurrency control topics.

\section*{Limitations}
While this position paper establishes a compelling conceptual mapping between MAS failures and classical concurrency anomalies, our framework primarily targets systems with explicitly shared mutable state, and may not generalize to MAS architectures that rely on implicit coordination or emergent communication. Additionally, the trade-offs among correctness, efficiency, and inference cost are discussed qualitatively; more rigorous empirical validation across diverse real-world workloads is left for future work.

\bibliography{refs}

@misc{icml2026sub,
      title={Position: Science is Collaborative—LLM for Science Should Be Too}, 
      author={Anonymous},
      year={2026},
      note={Filename in supplementary material: 52.pdf}, 
      howpublished ="Concurrent Submission to ICML"
}

@article{plume24,
  author       = {Si Liu and
                  Long Gu and
                  Hengfeng Wei and
                  David A. Basin},
  title        = {Plume: Efficient and Complete Black-Box Checking of Weak Isolation
                  Levels},
  journal      = {Proc. {ACM} Program. Lang.},
  volume       = {8},
  number       = {{OOPSLA2}},
  pages        = {876--904},
  year         = {2024},
  url          = {https://doi.org/10.1145/3689742},
  doi          = {10.1145/3689742},
  bibsource    = {dblp computer science bibliography, https://dblp.org}
}

@article{verystrong25,
  author       = {Zhiheng Cai and
                  Si Liu and
                  Hengfeng Wei and
                  Yuxing Chen and
                  Anqun Pan},
  title        = {Fast Verification of Strong Database Isolation},
  journal      = {Proc. {VLDB} Endow.},
  volume       = {19},
  number       = {4},
  pages        = {563--575},
  year         = {2025},
  url          = {https://www.vldb.org/pvldb/vol19/p563-wei.pdf},
  bibsource    = {dblp computer science bibliography, https://dblp.org}
}

@inproceedings{cobra20,
  author       = {Cheng Tan and
                  Changgeng Zhao and
                  Shuai Mu and
                  Michael Walfish},
  title        = {Cobra: Making Transactional Key-Value Stores Verifiably Serializable},
  booktitle    = {14th {USENIX} Symposium on Operating Systems Design and Implementation,
                  {OSDI} 2020, Virtual Event, November 4-6, 2020},
  pages        = {63--80},
  publisher    = {{USENIX} Association},
  year         = {2020},
  url          = {https://www.usenix.org/conference/osdi20/presentation/tan},
  bibsource    = {dblp computer science bibliography, https://dblp.org}
}

@inproceedings{brown2020language,
  author       = {Tom B. Brown and
                  Benjamin Mann and
                  Nick Ryder and
                  Melanie Subbiah and
                  Jared Kaplan and
                  Prafulla Dhariwal and
                  Arvind Neelakantan and
                  Pranav Shyam and
                  Girish Sastry and
                  Amanda Askell and
                  Sandhini Agarwal and
                  Ariel Herbert{-}Voss and
                  Gretchen Krueger and
                  Tom Henighan and
                  Rewon Child and
                  Aditya Ramesh and
                  Daniel M. Ziegler and
                  Jeffrey Wu and
                  Clemens Winter and
                  Christopher Hesse and
                  Mark Chen and
                  Eric Sigler and
                  Mateusz Litwin and
                  Scott Gray and
                  Benjamin Chess and
                  Jack Clark and
                  Christopher Berner and
                  Sam McCandlish and
                  Alec Radford and
                  Ilya Sutskever and
                  Dario Amodei},
  editor       = {Hugo Larochelle and
                  Marc'Aurelio Ranzato and
                  Raia Hadsell and
                  Maria{-}Florina Balcan and
                  Hsuan{-}Tien Lin},
  title        = {Language Models are Few-Shot Learners},
  booktitle    = {Advances in Neural Information Processing Systems 33: Annual Conference
                  on Neural Information Processing Systems 2020, NeurIPS 2020, December
                  6-12, 2020, virtual},
  year         = {2020},
  url          = {https://proceedings.neurips.cc/paper/2020/hash/1457c0d6bfcb4967418bfb8ac142f64a-Abstract.html},
  bibsource    = {dblp computer science bibliography, https://dblp.org}
}

@article{openai2024gpt4technicalreport,
  author       = {OpenAI},
  title        = {{GPT-4} Technical Report},
  journal      = {CoRR},
  volume       = {abs/2303.08774},
  year         = {2023},
  url          = {https://doi.org/10.48550/arXiv.2303.08774},
  doi          = {10.48550/ARXIV.2303.08774},
  eprinttype   = {arXiv},
  eprint       = {2303.08774},
  bibsource    = {dblp computer science bibliography, https://dblp.org}
}

@article{touvron2023llama,
  author       = {Hugo Touvron and
                  Louis Martin and
                  Kevin Stone and
                  Peter Albert and
                  Amjad Almahairi and
                  Yasmine Babaei and
                  Nikolay Bashlykov and
                  Soumya Batra and
                  Prajjwal Bhargava and
                  Shruti Bhosale and
                  Dan Bikel and
                  Lukas Blecher and
                  Cristian Canton{-}Ferrer and
                  Moya Chen and
                  Guillem Cucurull and
                  David Esiobu and
                  Jude Fernandes and
                  Jeremy Fu and
                  Wenyin Fu and
                  Brian Fuller and
                  Cynthia Gao and
                  Vedanuj Goswami and
                  Naman Goyal and
                  Anthony Hartshorn and
                  Saghar Hosseini and
                  Rui Hou and
                  Hakan Inan and
                  Marcin Kardas and
                  Viktor Kerkez and
                  Madian Khabsa and
                  Isabel Kloumann and
                  Artem Korenev and
                  Punit Singh Koura and
                  Marie{-}Anne Lachaux and
                  Thibaut Lavril and
                  Jenya Lee and
                  Diana Liskovich and
                  Yinghai Lu and
                  Yuning Mao and
                  Xavier Martinet and
                  Todor Mihaylov and
                  Pushkar Mishra and
                  Igor Molybog and
                  Yixin Nie and
                  Andrew Poulton and
                  Jeremy Reizenstein and
                  Rashi Rungta and
                  Kalyan Saladi and
                  Alan Schelten and
                  Ruan Silva and
                  Eric Michael Smith and
                  Ranjan Subramanian and
                  Xiaoqing Ellen Tan and
                  Binh Tang and
                  Ross Taylor and
                  Adina Williams and
                  Jian Xiang Kuan and
                  Puxin Xu and
                  Zheng Yan and
                  Iliyan Zarov and
                  Yuchen Zhang and
                  Angela Fan and
                  Melanie Kambadur and
                  Sharan Narang and
                  Aur{\'{e}}lien Rodriguez and
                  Robert Stojnic and
                  Sergey Edunov and
                  Thomas Scialom},
  title        = {Llama 2: Open Foundation and Fine-Tuned Chat Models},
  journal      = {CoRR},
  volume       = {abs/2307.09288},
  year         = {2023},
  url          = {https://doi.org/10.48550/arXiv.2307.09288},
  doi          = {10.48550/ARXIV.2307.09288},
  eprinttype   = {arXiv},
  eprint       = {2307.09288},
  bibsource    = {dblp computer science bibliography, https://dblp.org}
}

@inproceedings{yao2023react,
  author       = {Shunyu Yao and
                  Jeffrey Zhao and
                  Dian Yu and
                  Nan Du and
                  Izhak Shafran and
                  Karthik R. Narasimhan and
                  Yuan Cao},
  title        = {ReAct: Synergizing Reasoning and Acting in Language Models},
  booktitle    = {The Eleventh International Conference on Learning Representations,
                  {ICLR} 2023, Kigali, Rwanda, May 1-5, 2023},
  publisher    = {OpenReview.net},
  year         = {2023},
  url          = {https://openreview.net/forum?id=WE\_vluYUL-X},
  bibsource    = {dblp computer science bibliography, https://dblp.org}
}

@inproceedings{schick2023toolformer,
  author       = {Timo Schick and
                  Jane Dwivedi{-}Yu and
                  Roberto Dess{\`{\i}} and
                  Roberta Raileanu and
                  Maria Lomeli and
                  Eric Hambro and
                  Luke Zettlemoyer and
                  Nicola Cancedda and
                  Thomas Scialom},
  editor       = {Alice Oh and
                  Tristan Naumann and
                  Amir Globerson and
                  Kate Saenko and
                  Moritz Hardt and
                  Sergey Levine},
  title        = {Toolformer: Language Models Can Teach Themselves to Use Tools},
  booktitle    = {Advances in Neural Information Processing Systems 36: Annual Conference
                  on Neural Information Processing Systems 2023, NeurIPS 2023, New Orleans,
                  LA, USA, December 10 - 16, 2023},
  year         = {2023},
  url          = {http://papers.nips.cc/paper\_files/paper/2023/hash/d842425e4bf79ba039352da0f658a906-Abstract-Conference.html},
  bibsource    = {dblp computer science bibliography, https://dblp.org}
}

@article{mialon2023augmented,
  author       = {Gr{\'{e}}goire Mialon and
                  Roberto Dess{\`{\i}} and
                  Maria Lomeli and
                  Christoforos Nalmpantis and
                  Ramakanth Pasunuru and
                  Roberta Raileanu and
                  Baptiste Rozi{\`{e}}re and
                  Timo Schick and
                  Jane Dwivedi{-}Yu and
                  Asli Celikyilmaz and
                  Edouard Grave and
                  Yann LeCun and
                  Thomas Scialom},
  title        = {Augmented Language Models: a Survey},
  journal      = {Trans. Mach. Learn. Res.},
  volume       = {2023},
  year         = {2023},
  url          = {https://openreview.net/forum?id=jh7wH2AzKK},
  bibsource    = {dblp computer science bibliography, https://dblp.org}
}

@inproceedings{park2023generative,
  author       = {Joon Sung Park and
                  Joseph C. O'Brien and
                  Carrie Jun Cai and
                  Meredith Ringel Morris and
                  Percy Liang and
                  Michael S. Bernstein},
  editor       = {Sean Follmer and
                  Jeff Han and
                  J{\"{u}}rgen Steimle and
                  Nathalie Henry Riche},
  title        = {Generative Agents: Interactive Simulacra of Human Behavior},
  booktitle    = {Proceedings of the 36th Annual {ACM} Symposium on User Interface Software
                  and Technology, {UIST} 2023, San Francisco, CA, USA, 29 October 2023-
                  1 November 2023},
  pages        = {2:1--2:22},
  publisher    = {{ACM}},
  year         = {2023},
  url          = {https://doi.org/10.1145/3586183.3606763},
  doi          = {10.1145/3586183.3606763},
  bibsource    = {dblp computer science bibliography, https://dblp.org}
}

@inproceedings{li2023camel,
  author       = {Guohao Li and
                  Hasan Hammoud and
                  Hani Itani and
                  Dmitrii Khizbullin and
                  Bernard Ghanem},
  editor       = {Alice Oh and
                  Tristan Naumann and
                  Amir Globerson and
                  Kate Saenko and
                  Moritz Hardt and
                  Sergey Levine},
  title        = {{CAMEL:} Communicative Agents for "Mind" Exploration of Large Language
                  Model Society},
  booktitle    = {Advances in Neural Information Processing Systems 36: Annual Conference
                  on Neural Information Processing Systems 2023, NeurIPS 2023, New Orleans,
                  LA, USA, December 10 - 16, 2023},
  year         = {2023},
  url          = {http://papers.nips.cc/paper\_files/paper/2023/hash/a3621ee907def47c1b952ade25c67698-Abstract-Conference.html},
  bibsource    = {dblp computer science bibliography, https://dblp.org}
}

@inproceedings{hong2024metagpt,
  author       = {Sirui Hong and
                  Mingchen Zhuge and
                  Jonathan Chen and
                  Xiawu Zheng and
                  Yuheng Cheng and
                  Jinlin Wang and
                  Ceyao Zhang and
                  Zili Wang and
                  Steven Ka Shing Yau and
                  Zijuan Lin and
                  Liyang Zhou and
                  Chenyu Ran and
                  Lingfeng Xiao and
                  Chenglin Wu and
                  J{\"{u}}rgen Schmidhuber},
  title        = {MetaGPT: Meta Programming for {A} Multi-Agent Collaborative Framework},
  booktitle    = {The Twelfth International Conference on Learning Representations,
                  {ICLR} 2024, Vienna, Austria, May 7-11, 2024},
  publisher    = {OpenReview.net},
  year         = {2024},
  url          = {https://openreview.net/forum?id=VtmBAGCN7o},
  bibsource    = {dblp computer science bibliography, https://dblp.org}
}

@inproceedings{qian2024chatdev,
  author       = {Chen Qian and
                  Wei Liu and
                  Hongzhang Liu and
                  Nuo Chen and
                  Yufan Dang and
                  Jiahao Li and
                  Cheng Yang and
                  Weize Chen and
                  Yusheng Su and
                  Xin Cong and
                  Juyuan Xu and
                  Dahai Li and
                  Zhiyuan Liu and
                  Maosong Sun},
  editor       = {Lun{-}Wei Ku and
                  Andre Martins and
                  Vivek Srikumar},
  title        = {ChatDev: Communicative Agents for Software Development},
  booktitle    = {Proceedings of the 62nd Annual Meeting of the Association for Computational
                  Linguistics (Volume 1: Long Papers), {ACL} 2024, Bangkok, Thailand,
                  August 11-16, 2024},
  pages        = {15174--15186},
  publisher    = {Association for Computational Linguistics},
  year         = {2024},
  url          = {https://doi.org/10.18653/v1/2024.acl-long.810},
  doi          = {10.18653/V1/2024.ACL-LONG.810},
  bibsource    = {dblp computer science bibliography, https://dblp.org}
}

@article{wu2024autogen,
  author       = {Qingyun Wu and
                  Gagan Bansal and
                  Jieyu Zhang and
                  Yiran Wu and
                  Shaokun Zhang and
                  Erkang Zhu and
                  Beibin Li and
                  Li Jiang and
                  Xiaoyun Zhang and
                  Chi Wang},
  title        = {AutoGen: Enabling Next-Gen {LLM} Applications via Multi-Agent Conversation
                  Framework},
  journal      = {CoRR},
  volume       = {abs/2308.08155},
  year         = {2023},
  url          = {https://doi.org/10.48550/arXiv.2308.08155},
  doi          = {10.48550/ARXIV.2308.08155},
  eprinttype   = {arXiv},
  eprint       = {2308.08155},
  bibsource    = {dblp computer science bibliography, https://dblp.org}
}

@inproceedings{guo2024large,
  author       = {Taicheng Guo and
                  Xiuying Chen and
                  Yaqi Wang and
                  Ruidi Chang and
                  Shichao Pei and
                  Nitesh V. Chawla and
                  Olaf Wiest and
                  Xiangliang Zhang},
  title        = {Large Language Model Based Multi-agents: {A} Survey of Progress and
                  Challenges},
  booktitle    = {Proceedings of the Thirty-Third International Joint Conference on
                  Artificial Intelligence, {IJCAI} 2024, Jeju, South Korea, August 3-9,
                  2024},
  pages        = {8048--8057},
  publisher    = {ijcai.org},
  year         = {2024},
  url          = {https://www.ijcai.org/proceedings/2024/890},
  bibsource    = {dblp computer science bibliography, https://dblp.org}
}

@article{kim2025towards,
  author       = {Yubin Kim and
                  Ken Gu and
                  Chanwoo Park and
                  Chunjong Park and
                  Samuel Schmidgall and
                  A. Ali Heydari and
                  Yao Yan and
                  Zhihan Zhang and
                  Yuchen Zhuang and
                  Mark Malhotra and
                  Paul Pu Liang and
                  Hae Won Park and
                  Yuzhe Yang and
                  Xuhai Xu and
                  Yilun Du and
                  Shwetak N. Patel and
                  Tim Althoff and
                  Daniel McDuff and
                  Xin Liu},
  title        = {Towards a Science of Scaling Agent Systems},
  journal      = {CoRR},
  volume       = {abs/2512.08296},
  year         = {2025},
  url          = {https://doi.org/10.48550/arXiv.2512.08296},
  doi          = {10.48550/ARXIV.2512.08296},
  eprinttype   = {arXiv},
  eprint       = {2512.08296},
  bibsource    = {dblp computer science bibliography, https://dblp.org}
}

@article{li2024more,
  author       = {Junyou Li and
                  Qin Zhang and
                  Yangbin Yu and
                  Qiang Fu and
                  Deheng Ye},
  title        = {More Agents Is All You Need},
  journal      = {Trans. Mach. Learn. Res.},
  volume       = {2024},
  year         = {2024},
  url          = {https://openreview.net/forum?id=bgzUSZ8aeg},
  bibsource    = {dblp computer science bibliography, https://dblp.org}
}

@article{bernstein1981concurrency,
  author       = {Philip A. Bernstein and
                  Nathan Goodman},
  title        = {Concurrency Control in Distributed Database Systems},
  journal      = {{ACM} Comput. Surv.},
  volume       = {13},
  number       = {2},
  pages        = {185--221},
  year         = {1981},
  url          = {https://doi.org/10.1145/356842.356846},
  doi          = {10.1145/356842.356846},
  bibsource    = {dblp computer science bibliography, https://dblp.org}
}

@book{bernstein1987concurrency,
  author       = {Philip A. Bernstein and
                  Vassos Hadzilacos and
                  Nathan Goodman},
  title        = {Concurrency Control and Recovery in Database Systems},
  publisher    = {Addison-Wesley},
  year         = {1987},
  url          = {http://research.microsoft.com/en-us/people/philbe/ccontrol.aspx},
  isbn         = {0-201-10715-5},
  bibsource    = {dblp computer science bibliography, https://dblp.org}
}

@article{li2024survey,
  title={A survey on LLM-based multi-agent systems: workflow, infrastructure, and challenges},
  author={Li, Xinyi and Wang, Sai and Zeng, Siqi and Wu, Yu and Yang, Yi},
  journal={Vicinagearth},
  volume={1},
  number={1},
  pages={9},
  year={2024},
  publisher={Springer}
}

@article{chen2024survey,
  title={A survey on llm-based multi-agent system: Recent advances and new frontiers in application},
  author={Chen, Shuaihang and Liu, Yuanxing and Han, Wei and Zhang, Weinan and Liu, Ting},
  journal={arXiv preprint arXiv:2412.17481},
  year={2024}
}

@article{haerder1983principles,
  author       = {Theo H{\"{a}}rder and
                  Andreas Reuter},
  title        = {Principles of Transaction-Oriented Database Recovery},
  journal      = {{ACM} Comput. Surv.},
  volume       = {15},
  number       = {4},
  pages        = {287--317},
  year         = {1983},
  url          = {https://doi.org/10.1145/289.291},
  doi          = {10.1145/289.291},
  bibsource    = {dblp computer science bibliography, https://dblp.org}
}

@book{weikum2001transactional,
  author       = {Gerhard Weikum and
                  Gottfried Vossen},
  title        = {Transactional Information Systems: Theory, Algorithms, and the Practice
                  of Concurrency Control and Recovery},
  publisher    = {Morgan Kaufmann},
  year         = {2002},
  isbn         = {1-55860-508-8},
  bibsource    = {dblp computer science bibliography, https://dblp.org}
}

@techreport{adya1999weak,
author = {Adya, A.},
title = {Weak Consistency: A Generalized Theory and Optimistic Implementations for Distributed Transactions},
year = {1999},
institution = {Massachusetts Institute of Technology},
publisher = {Massachusetts Institute of Technology},
address = {USA}
}

@inproceedings{berenson1995critique,
  author       = {Hal Berenson and
                  Philip A. Bernstein and
                  Jim Gray and
                  Jim Melton and
                  Elizabeth J. O'Neil and
                  Patrick E. O'Neil},
  editor       = {Michael J. Carey and
                  Donovan A. Schneider},
  title        = {A Critique of {ANSI} {SQL} Isolation Levels},
  booktitle    = {Proceedings of the 1995 {ACM} {SIGMOD} International Conference on
                  Management of Data, San Jose, California, USA, May 22-25, 1995},
  pages        = {1--10},
  publisher    = {{ACM} Press},
  year         = {1995},
  url          = {https://doi.org/10.1145/223784.223785},
  doi          = {10.1145/223784.223785},
  bibsource    = {dblp computer science bibliography, https://dblp.org}
}

@article{eswaran1976notions,
  author       = {Kapali P. Eswaran and
                  Jim Gray and
                  Raymond A. Lorie and
                  Irving L. Traiger},
  title        = {The Notions of Consistency and Predicate Locks in a Database System},
  journal      = {Commun. {ACM}},
  volume       = {19},
  number       = {11},
  pages        = {624--633},
  year         = {1976},
  url          = {https://doi.org/10.1145/360363.360369},
  doi          = {10.1145/360363.360369},
  bibsource    = {dblp computer science bibliography, https://dblp.org}
}

@article{kung1981optimistic,
  author       = {H. T. Kung and
                  John T. Robinson},
  title        = {On Optimistic Methods for Concurrency Control},
  journal      = {{ACM} Trans. Database Syst.},
  volume       = {6},
  number       = {2},
  pages        = {213--226},
  year         = {1981},
  url          = {https://doi.org/10.1145/319566.319567},
  doi          = {10.1145/319566.319567},
  bibsource    = {dblp computer science bibliography, https://dblp.org}
}

@phdthesis{reed1978naming,
  author       = {David P. Reed},
  title        = {Naming and synchronization in a decentralized computer system},
  school       = {Massachusetts Institute of Technology, Cambridge, MA, {USA}},
  year         = {1978},
  url          = {https://hdl.handle.net/1721.1/16279},
  bibsource    = {dblp computer science bibliography, https://dblp.org}
}

@article{papadimitriou1979serializability,
  author       = {Christos H. Papadimitriou},
  title        = {The serializability of concurrent database updates},
  journal      = {J. {ACM}},
  volume       = {26},
  number       = {4},
  pages        = {631--653},
  year         = {1979},
  url          = {https://doi.org/10.1145/322154.322158},
  doi          = {10.1145/322154.322158},
  bibsource    = {dblp computer science bibliography, https://dblp.org}
}

@inproceedings{fan2022minedojo,
  author       = {Linxi Fan and
                  Guanzhi Wang and
                  Yunfan Jiang and
                  Ajay Mandlekar and
                  Yuncong Yang and
                  Haoyi Zhu and
                  Andrew Tang and
                  De{-}An Huang and
                  Yuke Zhu and
                  Anima Anandkumar},
  editor       = {Sanmi Koyejo and
                  S. Mohamed and
                  A. Agarwal and
                  Danielle Belgrave and
                  K. Cho and
                  A. Oh},
  title        = {MineDojo: Building Open-Ended Embodied Agents with Internet-Scale
                  Knowledge},
  booktitle    = {Advances in Neural Information Processing Systems 35: Annual Conference
                  on Neural Information Processing Systems 2022, NeurIPS 2022, New Orleans,
                  LA, USA, November 28 - December 9, 2022},
  year         = {2022},
  url          = {http://papers.nips.cc/paper\_files/paper/2022/hash/74a67268c5cc5910f64938cac4526a90-Abstract-Datasets\_and\_Benchmarks.html},
  bibsource    = {dblp computer science bibliography, https://dblp.org}
}

@article{wang2023voyager,
  author       = {Guanzhi Wang and
                  Yuqi Xie and
                  Yunfan Jiang and
                  Ajay Mandlekar and
                  Chaowei Xiao and
                  Yuke Zhu and
                  Linxi Fan and
                  Anima Anandkumar},
  title        = {Voyager: An Open-Ended Embodied Agent with Large Language Models},
  journal      = {Trans. Mach. Learn. Res.},
  volume       = {2024},
  year         = {2024},
  url          = {https://openreview.net/forum?id=ehfRiF0R3a},
  bibsource    = {dblp computer science bibliography, https://dblp.org}
}

@inproceedings{dong2024villageragent,
  author       = {Yubo Dong and
                  Xukun Zhu and
                  Zhengzhe Pan and
                  Linchao Zhu and
                  Yi Yang},
  editor       = {Lun{-}Wei Ku and
                  Andre Martins and
                  Vivek Srikumar},
  title        = {VillagerAgent: {A} Graph-Based Multi-Agent Framework for Coordinating
                  Complex Task Dependencies in Minecraft},
  booktitle    = {Findings of the Association for Computational Linguistics, {ACL} 2024,
                  Bangkok, Thailand and virtual meeting, August 11-16, 2024},
  series       = {Findings of {ACL}},
  pages        = {16290--16314},
  publisher    = {Association for Computational Linguistics},
  year         = {2024},
  url          = {https://doi.org/10.18653/v1/2024.findings-acl.964},
  doi          = {10.18653/V1/2024.FINDINGS-ACL.964},
  bibsource    = {dblp computer science bibliography, https://dblp.org}
}

@article{fekete2005making,
  author       = {Alan D. Fekete and
                  Dimitrios Liarokapis and
                  Elizabeth J. O'Neil and
                  Patrick E. O'Neil and
                  Dennis E. Shasha},
  title        = {Making snapshot isolation serializable},
  journal      = {{ACM} Trans. Database Syst.},
  volume       = {30},
  number       = {2},
  pages        = {492--528},
  year         = {2005},
  url          = {https://doi.org/10.1145/1071610.1071615},
  doi          = {10.1145/1071610.1071615},
  bibsource    = {dblp computer science bibliography, https://dblp.org}
}

@inproceedings{gray1976granularity,
  author       = {Jim Gray and
                  Raymond A. Lorie and
                  Gianfranco R. Putzolu and
                  Irving L. Traiger},
  editor       = {G. M. Nijssen},
  title        = {Granularity of Locks and Degrees of Consistency in a Shared Data Base},
  booktitle    = {Modelling in Data Base Management Systems, Proceeding of the {IFIP}
                  Working Conference on Modelling in Data Base Management Systems, Freudenstadt,
                  Germany, January 5-8, 1976},
  pages        = {365--394},
  publisher    = {North-Holland},
  year         = {1976},
  bibsource    = {dblp computer science bibliography, https://dblp.org}
}

@inproceedings{lowe2017multi,
  author       = {Ryan Lowe and
                  Yi Wu and
                  Aviv Tamar and
                  Jean Harb and
                  Pieter Abbeel and
                  Igor Mordatch},
  editor       = {Isabelle Guyon and
                  Ulrike von Luxburg and
                  Samy Bengio and
                  Hanna M. Wallach and
                  Rob Fergus and
                  S. V. N. Vishwanathan and
                  Roman Garnett},
  title        = {Multi-Agent Actor-Critic for Mixed Cooperative-Competitive Environments},
  booktitle    = {Advances in Neural Information Processing Systems 30: Annual Conference
                  on Neural Information Processing Systems 2017, December 4-9, 2017,
                  Long Beach, CA, {USA}},
  pages        = {6379--6390},
  year         = {2017},
  url          = {https://proceedings.neurips.cc/paper/2017/hash/68a9750337a418a86fe06c1991a1d64c-Abstract.html},
  bibsource    = {dblp computer science bibliography, https://dblp.org}
}

@inproceedings{foerster2018learning,
  author       = {Jakob N. Foerster and
                  Richard Y. Chen and
                  Maruan Al{-}Shedivat and
                  Shimon Whiteson and
                  Pieter Abbeel and
                  Igor Mordatch},
  editor       = {Elisabeth Andr{\'{e}} and
                  Sven Koenig and
                  Mehdi Dastani and
                  Gita Sukthankar},
  title        = {Learning with Opponent-Learning Awareness},
  booktitle    = {Proceedings of the 17th International Conference on Autonomous Agents
                  and MultiAgent Systems, {AAMAS} 2018, Stockholm, Sweden, July 10-15,
                  2018},
  pages        = {122--130},
  publisher    = {International Foundation for Autonomous Agents and Multiagent Systems
                  Richland, SC, {USA} / {ACM}},
  year         = {2018},
  url          = {http://dl.acm.org/citation.cfm?id=3237408},
  bibsource    = {dblp computer science bibliography, https://dblp.org}
}

@inproceedings{yu2022surprising,
  author       = {Chao Yu and
                  Akash Velu and
                  Eugene Vinitsky and
                  Jiaxuan Gao and
                  Yu Wang and
                  Alexandre M. Bayen and
                  Yi Wu},
  editor       = {Sanmi Koyejo and
                  S. Mohamed and
                  A. Agarwal and
                  Danielle Belgrave and
                  K. Cho and
                  A. Oh},
  title        = {The Surprising Effectiveness of {PPO} in Cooperative Multi-Agent Games},
  booktitle    = {Advances in Neural Information Processing Systems 35: Annual Conference
                  on Neural Information Processing Systems 2022, NeurIPS 2022, New Orleans,
                  LA, USA, November 28 - December 9, 2022},
  year         = {2022},
  url          = {http://papers.nips.cc/paper\_files/paper/2022/hash/9c1535a02f0ce079433344e14d910597-Abstract-Datasets\_and\_Benchmarks.html},
  bibsource    = {dblp computer science bibliography, https://dblp.org}
}

@article{hernandez2017survey,
  author       = {Pablo Hernandez{-}Leal and
                  Michael Kaisers and
                  Tim Baarslag and
                  Enrique Munoz de Cote},
  title        = {A Survey of Learning in Multiagent Environments: Dealing with Non-Stationarity},
  journal      = {CoRR},
  volume       = {abs/1707.09183},
  year         = {2017},
  url          = {http://arxiv.org/abs/1707.09183},
  eprinttype   = {arXiv},
  eprint       = {1707.09183},
  bibsource    = {dblp computer science bibliography, https://dblp.org}
}

@inproceedings{yu2022orca,
  author       = {Gyeong{-}In Yu and
                  Joo Seong Jeong and
                  Geon{-}Woo Kim and
                  Soojeong Kim and
                  Byung{-}Gon Chun},
  editor       = {Marcos K. Aguilera and
                  Hakim Weatherspoon},
  title        = {Orca: {A} Distributed Serving System for Transformer-Based Generative
                  Models},
  booktitle    = {16th {USENIX} Symposium on Operating Systems Design and Implementation,
                  {OSDI} 2022, Carlsbad, CA, USA, July 11-13, 2022},
  pages        = {521--538},
  publisher    = {{USENIX} Association},
  year         = {2022},
  url          = {https://www.usenix.org/conference/osdi22/presentation/yu},
  bibsource    = {dblp computer science bibliography, https://dblp.org}
}

@inproceedings{kwon2023efficient,
  author       = {Woosuk Kwon and
                  Zhuohan Li and
                  Siyuan Zhuang and
                  Ying Sheng and
                  Lianmin Zheng and
                  Cody Hao Yu and
                  Joseph Gonzalez and
                  Hao Zhang and
                  Ion Stoica},
  editor       = {Jason Flinn and
                  Margo I. Seltzer and
                  Peter Druschel and
                  Antoine Kaufmann and
                  Jonathan Mace},
  title        = {Efficient Memory Management for Large Language Model Serving with
                  PagedAttention},
  booktitle    = {Proceedings of the 29th Symposium on Operating Systems Principles,
                  {SOSP} 2023, Koblenz, Germany, October 23-26, 2023},
  pages        = {611--626},
  publisher    = {{ACM}},
  year         = {2023},
  url          = {https://doi.org/10.1145/3600006.3613165},
  doi          = {10.1145/3600006.3613165},
  bibsource    = {dblp computer science bibliography, https://dblp.org}
}

@inproceedings{leviathan2023fast,
  author       = {Yaniv Leviathan and
                  Matan Kalman and
                  Yossi Matias},
  editor       = {Andreas Krause and
                  Emma Brunskill and
                  Kyunghyun Cho and
                  Barbara Engelhardt and
                  Sivan Sabato and
                  Jonathan Scarlett},
  title        = {Fast Inference from Transformers via Speculative Decoding},
  booktitle    = {International Conference on Machine Learning, {ICML} 2023, 23-29 July
                  2023, Honolulu, Hawaii, {USA}},
  series       = {Proceedings of Machine Learning Research},
  pages        = {19274--19286},
  publisher    = {{PMLR}},
  year         = {2023},
  url          = {https://proceedings.mlr.press/v202/leviathan23a.html},
  bibsource    = {dblp computer science bibliography, https://dblp.org}
}

@article{frantar2022gptq,
  author       = {Elias Frantar and
                  Saleh Ashkboos and
                  Torsten Hoefler and
                  Dan Alistarh},
  title        = {{GPTQ:} Accurate Post-Training Quantization for Generative Pre-trained
                  Transformers},
  journal      = {CoRR},
  volume       = {abs/2210.17323},
  year         = {2022},
  url          = {https://doi.org/10.48550/arXiv.2210.17323},
  doi          = {10.48550/ARXIV.2210.17323},
  eprinttype   = {arXiv},
  eprint       = {2210.17323},
  bibsource    = {dblp computer science bibliography, https://dblp.org}
}

@inproceedings{zheng2024sglang,
  author       = {Lianmin Zheng and
                  Liangsheng Yin and
                  Zhiqiang Xie and
                  Chuyue Sun and
                  Jeff Huang and
                  Cody Hao Yu and
                  Shiyi Cao and
                  Christos Kozyrakis and
                  Ion Stoica and
                  Joseph E. Gonzalez and
                  Clark W. Barrett and
                  Ying Sheng},
  editor       = {Amir Globersons and
                  Lester Mackey and
                  Danielle Belgrave and
                  Angela Fan and
                  Ulrich Paquet and
                  Jakub M. Tomczak and
                  Cheng Zhang},
  title        = {SGLang: Efficient Execution of Structured Language Model Programs},
  booktitle    = {Advances in Neural Information Processing Systems 38: Annual Conference
                  on Neural Information Processing Systems 2024, NeurIPS 2024, Vancouver,
                  BC, Canada, December 10 - 15, 2024},
  year         = {2024},
  url          = {http://papers.nips.cc/paper\_files/paper/2024/hash/724be4472168f31ba1c9ac630f15dec8-Abstract-Conference.html},
  bibsource    = {dblp computer science bibliography, https://dblp.org}
}

@misc{pinecone2023,
  title={Pinecone: The vector database for machine learning},
  author={{Pinecone Systems}},
  year={2023},
  howpublished={\url{https://www.pinecone.io/}}
}

@misc{weaviate2023,
  title={Weaviate: The {AI}-native vector database},
  author={{Weaviate}},
  year={2023},
  howpublished={\url{https://weaviate.io/}}
}

@inproceedings{liu2024agentbench,
  author       = {Xiao Liu and
                  Hao Yu and
                  Hanchen Zhang and
                  Yifan Xu and
                  Xuanyu Lei and
                  Hanyu Lai and
                  Yu Gu and
                  Hangliang Ding and
                  Kaiwen Men and
                  Kejuan Yang and
                  Shudan Zhang and
                  Xiang Deng and
                  Aohan Zeng and
                  Zhengxiao Du and
                  Chenhui Zhang and
                  Sheng Shen and
                  Tianjun Zhang and
                  Yu Su and
                  Huan Sun and
                  Minlie Huang and
                  Yuxiao Dong and
                  Jie Tang},
  title        = {AgentBench: Evaluating LLMs as Agents},
  booktitle    = {The Twelfth International Conference on Learning Representations,
                  {ICLR} 2024, Vienna, Austria, May 7-11, 2024},
  publisher    = {OpenReview.net},
  year         = {2024},
  url          = {https://openreview.net/forum?id=zAdUB0aCTQ},
  bibsource    = {dblp computer science bibliography, https://dblp.org}
}

@article{cemri2025multi,
  author       = {Mert Cemri and
                  Melissa Z. Pan and
                  Shuyi Yang and
                  Lakshya A. Agrawal and
                  Bhavya Chopra and
                  Rishabh Tiwari and
                  Kurt Keutzer and
                  Aditya G. Parameswaran and
                  Dan Klein and
                  Kannan Ramchandran and
                  Matei Zaharia and
                  Joseph E. Gonzalez and
                  Ion Stoica},
  title        = {Why Do Multi-Agent {LLM} Systems Fail?},
  journal      = {CoRR},
  volume       = {abs/2503.13657},
  year         = {2025},
  url          = {https://doi.org/10.48550/arXiv.2503.13657},
  doi          = {10.48550/ARXIV.2503.13657},
  eprinttype   = {arXiv},
  eprint       = {2503.13657},
  bibsource    = {dblp computer science bibliography, https://dblp.org}
}

@article{cahill2009serializable,
  author       = {Michael J. Cahill and
                  Uwe R{\"{o}}hm and
                  Alan D. Fekete},
  title        = {Serializable isolation for snapshot databases},
  journal      = {{ACM} Trans. Database Syst.},
  volume       = {34},
  number       = {4},
  pages        = {20:1--20:42},
  year         = {2009},
  url          = {https://doi.org/10.1145/1620585.1620587},
  doi          = {10.1145/1620585.1620587},
  bibsource    = {dblp computer science bibliography, https://dblp.org}
}

@inproceedings{jimenez2024swe,
  author       = {Carlos E. Jimenez and
                  John Yang and
                  Alexander Wettig and
                  Shunyu Yao and
                  Kexin Pei and
                  Ofir Press and
                  Karthik R. Narasimhan},
  title        = {SWE-bench: Can Language Models Resolve Real-world Github Issues?},
  booktitle    = {The Twelfth International Conference on Learning Representations,
                  {ICLR} 2024, Vienna, Austria, May 7-11, 2024},
  publisher    = {OpenReview.net},
  year         = {2024},
  url          = {https://openreview.net/forum?id=VTF8yNQM66},
  bibsource    = {dblp computer science bibliography, https://dblp.org}
}

@inproceedings{zhou2024webarena,
  author       = {Shuyan Zhou and
                  Frank F. Xu and
                  Hao Zhu and
                  Xuhui Zhou and
                  Robert Lo and
                  Abishek Sridhar and
                  Xianyi Cheng and
                  Tianyue Ou and
                  Yonatan Bisk and
                  Daniel Fried and
                  Uri Alon and
                  Graham Neubig},
  title        = {WebArena: {A} Realistic Web Environment for Building Autonomous Agents},
  booktitle    = {The Twelfth International Conference on Learning Representations,
                  {ICLR} 2024, Vienna, Austria, May 7-11, 2024},
  publisher    = {OpenReview.net},
  year         = {2024},
  url          = {https://openreview.net/forum?id=oKn9c6ytLx},
  bibsource    = {dblp computer science bibliography, https://dblp.org}
}

@misc{atil2024non,
      title={Non-Determinism of "Deterministic" LLM Settings}, 
      author={Berk Atil and Sarp Aykent and Alexa Chittams and Lisheng Fu and Rebecca J. Passonneau and Evan Radcliffe and Guru Rajan Rajagopal and Adam Sloan and Tomasz Tudrej and Ferhan Ture and Zhe Wu and Lixinyu Xu and Breck Baldwin},
      year={2024},
      eprint={2408.04667},
      archivePrefix={arXiv},
      primaryClass={cs.CL},
      url={https://arxiv.org/abs/2408.04667}, 
}

@inproceedings{kahng2024llm,
  author       = {Minsuk Kahng and
                  Ian Tenney and
                  Mahima Pushkarna and
                  Michael Xieyang Liu and
                  James Wexler and
                  Emily Reif and
                  Krystal Kallarackal and
                  Minsuk Chang and
                  Michael Terry and
                  Lucas Dixon},
  editor       = {Florian 'Floyd' Mueller and
                  Penny Kyburz and
                  Julie R. Williamson and
                  Corina Sas},
  title        = {{LLM} Comparator: Visual Analytics for Side-by-Side Evaluation of
                  Large Language Models},
  booktitle    = {Extended Abstracts of the {CHI} Conference on Human Factors in Computing
                  Systems, {CHI} {EA} 2024, Honolulu, HI, USA, May 11-16, 2024},
  pages        = {216:1--216:7},
  publisher    = {{ACM}},
  year         = {2024},
  url          = {https://doi.org/10.1145/3613905.3650755},
  doi          = {10.1145/3613905.3650755},
  bibsource    = {dblp computer science bibliography, https://dblp.org}
}

@inproceedings{shavit1995software,
  author       = {Nir Shavit and
                  Dan Touitou},
  editor       = {James H. Anderson},
  title        = {Software Transactional Memory},
  booktitle    = {Proceedings of the Fourteenth Annual {ACM} Symposium on Principles
                  of Distributed Computing, Ottawa, Ontario, Canada, August 20-23, 1995},
  pages        = {204--213},
  publisher    = {{ACM}},
  year         = {1995},
  url          = {https://doi.org/10.1145/224964.224987},
  doi          = {10.1145/224964.224987},
  bibsource    = {dblp computer science bibliography, https://dblp.org}
}

@inproceedings{harris2005composable,
  author       = {Tim Harris and
                  Simon Marlow and
                  Simon L. Peyton Jones and
                  Maurice Herlihy},
  editor       = {Keshav Pingali and
                  Katherine A. Yelick and
                  Andrew S. Grimshaw},
  title        = {Composable memory transactions},
  booktitle    = {Proceedings of the {ACM} {SIGPLAN} Symposium on Principles and Practice
                  of Parallel Programming, {PPOPP} 2005, June 15-17, 2005, Chicago,
                  IL, {USA}},
  pages        = {48--60},
  publisher    = {{ACM}},
  year         = {2005},
  url          = {https://doi.org/10.1145/1065944.1065952},
  doi          = {10.1145/1065944.1065952},
  bibsource    = {dblp computer science bibliography, https://dblp.org}
}

@article{zhang2021multi,
  author       = {Kaiqing Zhang and
                  Zhuoran Yang and
                  Tamer Basar},
  title        = {Multi-Agent Reinforcement Learning: {A} Selective Overview of Theories
                  and Algorithms},
  journal      = {CoRR},
  volume       = {abs/1911.10635},
  year         = {2019},
  url          = {http://arxiv.org/abs/1911.10635},
  eprinttype   = {arXiv},
  eprint       = {1911.10635},
  bibsource    = {dblp computer science bibliography, https://dblp.org}
}

@article{huang2025evogit,
  author       = {Beichen Huang and
                  Ran Cheng and
                  Kay Chen Tan},
  title        = {EvoGit: Decentralized Code Evolution via Git-Based Multi-Agent Collaboration},
  journal      = {CoRR},
  volume       = {abs/2506.02049},
  year         = {2025},
  url          = {https://doi.org/10.48550/arXiv.2506.02049},
  doi          = {10.48550/ARXIV.2506.02049},
  eprinttype   = {arXiv},
  eprint       = {2506.02049},
  bibsource    = {dblp computer science bibliography, https://dblp.org}
}

@article{wu2025git,
  author       = {Junde Wu},
  title        = {Git Context Controller: Manage the Context of LLM-based Agents like
                  Git},
  journal      = {CoRR},
  volume       = {abs/2508.00031},
  year         = {2025},
  url          = {https://doi.org/10.48550/arXiv.2508.00031},
  doi          = {10.48550/ARXIV.2508.00031},
  eprinttype   = {arXiv},
  eprint       = {2508.00031},
  bibsource    = {dblp computer science bibliography, https://dblp.org}
}

@inproceedings{tao2024magis,
  author       = {Wei Tao and
                  Yucheng Zhou and
                  Yanlin Wang and
                  Wenqiang Zhang and
                  Hongyu Zhang and
                  Yu Cheng},
  editor       = {Amir Globersons and
                  Lester Mackey and
                  Danielle Belgrave and
                  Angela Fan and
                  Ulrich Paquet and
                  Jakub M. Tomczak and
                  Cheng Zhang},
  title        = {{MAGIS:} LLM-Based Multi-Agent Framework for GitHub Issue Resolution},
  booktitle    = {Advances in Neural Information Processing Systems 38: Annual Conference
                  on Neural Information Processing Systems 2024, NeurIPS 2024, Vancouver,
                  BC, Canada, December 10 - 15, 2024},
  year         = {2024},
  url          = {http://papers.nips.cc/paper\_files/paper/2024/hash/5d1f02132ef51602adf07000ca5b6138-Abstract-Conference.html},
  bibsource    = {dblp computer science bibliography, https://dblp.org}
}

@inproceedings{shapiro2011conflict,
  author       = {Marc Shapiro and
                  Nuno M. Pregui{\c{c}}a and
                  Carlos Baquero and
                  Marek Zawirski},
  editor       = {Xavier D{\'{e}}fago and
                  Franck Petit and
                  Vincent Villain},
  title        = {Conflict-Free Replicated Data Types},
  booktitle    = {Stabilization, Safety, and Security of Distributed Systems - 13th
                  International Symposium, {SSS} 2011, Grenoble, France, October 10-12,
                  2011. Proceedings},
  series       = {Lecture Notes in Computer Science},
  pages        = {386--400},
  publisher    = {Springer},
  year         = {2011},
  url          = {https://doi.org/10.1007/978-3-642-24550-3\_29},
  doi          = {10.1007/978-3-642-24550-3\_29},
  bibsource    = {dblp computer science bibliography, https://dblp.org}
}

@article{geng2026effective,
  author       = {Jiayi Geng and
                  Graham Neubig},
  title        = {Effective Strategies for Asynchronous Software Engineering Agents},
  journal      = {CoRR},
  volume       = {abs/2603.21489},
  year         = {2026},
  url          = {https://doi.org/10.48550/arXiv.2603.21489},
  doi          = {10.48550/ARXIV.2603.21489},
  eprinttype   = {arXiv},
  eprint       = {2603.21489},
  bibsource    = {dblp computer science bibliography, https://dblp.org}
}

@article{zhang2026silo,
  author       = {Yuzhe Zhang and
                  Feiran Liu and
                  Yi Shan and
                  Xinyi Huang and
                  Xin Yang and
                  Yueqi Zhu and
                  Xuxin Cheng and
                  Cao Liu and
                  Ke Zeng and
                  Terry Jingchen Zhang and
                  Wenyuan Jiang},
  title        = {Silo-Bench: {A} Scalable Environment for Evaluating Distributed Coordination
                  in Multi-Agent {LLM} Systems},
  journal      = {CoRR},
  volume       = {abs/2603.01045},
  year         = {2026},
  url          = {https://doi.org/10.48550/arXiv.2603.01045},
  doi          = {10.48550/ARXIV.2603.01045},
  eprinttype   = {arXiv},
  eprint       = {2603.01045},
  bibsource    = {dblp computer science bibliography, https://dblp.org}
}

@article{chen2024coder,
  author       = {Dong Chen and
                  Shaoxin Lin and
                  Muhan Zeng and
                  Daoguang Zan and
                  Jian{-}Gang Wang and
                  Anton Cheshkov and
                  Jun Sun and
                  Hao Yu and
                  Guoliang Dong and
                  Artem Aliev and
                  Jie Wang and
                  Xiao Cheng and
                  Guangtai Liang and
                  Yuchi Ma and
                  Pan Bian and
                  Tao Xie and
                  Qianxiang Wang},
  title        = {CodeR: Issue Resolving with Multi-Agent and Task Graphs},
  journal      = {CoRR},
  volume       = {abs/2406.01304},
  year         = {2024},
  url          = {https://doi.org/10.48550/arXiv.2406.01304},
  doi          = {10.48550/ARXIV.2406.01304},
  eprinttype   = {arXiv},
  eprint       = {2406.01304},
  bibsource    = {dblp computer science bibliography, https://dblp.org}
}

@article{chang2025sagallm,
  author       = {Edward Y. Chang and
                  Longling Geng},
  title        = {SagaLLM: Context Management, Validation, and Transaction Guarantees
                  for Multi-Agent {LLM} Planning},
  journal      = {Proc. {VLDB} Endow.},
  volume       = {18},
  number       = {12},
  pages        = {4874--4886},
  year         = {2025},
  url          = {https://www.vldb.org/pvldb/vol18/p4874-chang.pdf},
  doi          = {10.14778/3750601.3750611},
  bibsource    = {dblp computer science bibliography, https://dblp.org}
}

@article{wang2024megaagent,
  author       = {Qian Wang and
                  Tianyu Wang and
                  Qinbin Li and
                  Jingsheng Liang and
                  Bingsheng He},
  title        = {MegaAgent: {A} Practical Framework for Autonomous Cooperation in Large-Scale
                  {LLM} Agent Systems},
  journal      = {CoRR},
  volume       = {abs/2408.09955},
  year         = {2024},
  url          = {https://doi.org/10.48550/arXiv.2408.09955},
  doi          = {10.48550/ARXIV.2408.09955},
  eprinttype   = {arXiv},
  eprint       = {2408.09955},
  bibsource    = {dblp computer science bibliography, https://dblp.org}
}

@inproceedings{stone2010adhoc,
  author       = {Peter Stone and
                  Gal A. Kaminka and
                  Sarit Kraus and
                  Jeffrey S. Rosenschein},
  editor       = {Maria Fox and
                  David Poole},
  title        = {Ad Hoc Autonomous Agent Teams: Collaboration without Pre-Coordination},
  booktitle    = {Proceedings of the Twenty-Fourth {AAAI} Conference on Artificial Intelligence,
                  {AAAI} 2010, Atlanta, Georgia, USA, July 11-15, 2010},
  pages        = {1504--1509},
  publisher    = {{AAAI} Press},
  year         = {2010},
  url          = {https://doi.org/10.1609/aaai.v24i1.7529},
  doi          = {10.1609/AAAI.V24I1.7529},
  bibsource    = {dblp computer science bibliography, https://dblp.org}
}

@inproceedings{barrett2015cooperating,
  author       = {Samuel Barrett and
                  Peter Stone},
  editor       = {Blai Bonet and
                  Sven Koenig},
  title        = {Cooperating with Unknown Teammates in Complex Domains: {A} Robot Soccer
                  Case Study of Ad Hoc Teamwork},
  booktitle    = {Proceedings of the Twenty-Ninth {AAAI} Conference on Artificial Intelligence,
                  January 25-30, 2015, Austin, Texas, {USA}},
  pages        = {2010--2016},
  publisher    = {{AAAI} Press},
  year         = {2015},
  url          = {https://doi.org/10.1609/aaai.v29i1.9428},
  doi          = {10.1609/AAAI.V29I1.9428},
  bibsource    = {dblp computer science bibliography, https://dblp.org}
}

@article{kleppmann2017json,
  author       = {Martin Kleppmann and
                  Alastair R. Beresford},
  title        = {A Conflict-Free Replicated {JSON} Datatype},
  journal      = {{IEEE} Trans. Parallel Distributed Syst.},
  volume       = {28},
  number       = {10},
  pages        = {2733--2746},
  year         = {2017},
  url          = {https://doi.org/10.1109/TPDS.2017.2697382},
  doi          = {10.1109/TPDS.2017.2697382},
  bibsource    = {dblp computer science bibliography, https://dblp.org}
}

@article{tran2025multiagent,
  author       = {Khanh{-}Tung Tran and
                  Dung Dao and
                  Minh{-}Duong Nguyen and
                  Quoc{-}Viet Pham and
                  Barry O'Sullivan and
                  Hoang D. Nguyen},
  title        = {Multi-Agent Collaboration Mechanisms: {A} Survey of LLMs},
  journal      = {CoRR},
  volume       = {abs/2501.06322},
  year         = {2025},
  url          = {https://doi.org/10.48550/arXiv.2501.06322},
  doi          = {10.48550/ARXIV.2501.06322},
  eprinttype   = {arXiv},
  eprint       = {2501.06322},
  bibsource    = {dblp computer science bibliography, https://dblp.org}
}
\bibliographystyle{icml2026}

\newpage
\appendix
\onecolumn

\section{LLM Usage}
We used large language models (LLMs) solely for writing assistance, including grammar checking, language polishing, and phrasing refinement. All technical content, experimental design, and conclusions are entirely the work of the authors.


\end{document}